%% file: iclr2026_conference.tex
\documentclass{article} 
\usepackage{iclr2026_conference,times}

\input{math_commands.tex}

\usepackage{hyperref}
\usepackage{algorithm}
\usepackage{algorithmic}
\usepackage{multirow}
\usepackage{graphicx}
\usepackage{booktabs}
\usepackage{amssymb}
\usepackage{pifont}
\usepackage{array}
\usepackage{longtable}
\usepackage{booktabs}
\usepackage{multirow}
\usepackage{xcolor}
\usepackage{amsmath}
\usepackage{tikz}
\usepackage{minted}
\usepackage{wrapfig}     
\usepackage{caption}     
\usepackage{subcaption}  
\usepackage{soul}
\usepackage[none]{hyphenat}

\definecolor{green}{rgb}{0.0, 0.6, 0.0}
\definecolor{lightgreen}{HTML}{2DC75C}
\definecolor{lightblue}{HTML}{4DA8DA}
\definecolor{lightpurple}{HTML}{9A8AD2}
\definecolor{lightgray}{HTML}{F2F2F2}
\definecolor{program}{HTML}{f6dfde}
\definecolor{perceptual}{HTML}{ffe2c9}
\definecolor{symbolic}{HTML}{c6dcff}

\usepackage[most]{tcolorbox}
\tcbset{
  colback=blue!5!white,   
  colframe=blue!50!black, 
  boxrule=0.4pt,         
  arc=2pt,               
  left=6pt,right=6pt,    
  top=6pt,bottom=6pt,
}
\tcbset{
  inlinebox/.style={
    enhanced,
    on line,            
    tcbox raise base,   
    boxrule=0pt,
    frame hidden,
    rounded corners,
    colback=#1!30,
    boxsep=0.5pt,
    left=2pt,right=2pt,top=0pt,bottom=0pt,
  }
}

\newtcolorbox{calloutbox}[1][]{
  enhanced,
  breakable,             
  width=\columnwidth,    
  title=\textbf{#1},
  fonttitle=\bfseries,
}

\newtcolorbox{promptbox}[1][]{
  enhanced,
  width=\columnwidth,    
  colback=lightgray,     
  colframe=black,        
  boxrule=1.4pt,         
  arc=4pt,               
  outer arc=4pt,
  left=6pt,right=6pt,top=6pt,bottom=6pt,
  #1                     
}

\title{\texorpdfstring{\raisebox{-0.2em}{\includegraphics[height=1.1em]{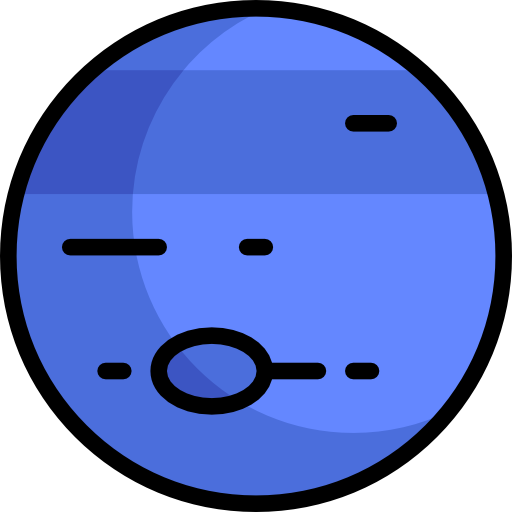}}}{} NePTune: A Neuro-Pythonic Framework for Tunable Compositional Reasoning on Vision-Language}

\iclrfinalcopy
\author{Danial Kamali \quad  Parisa Kordjamshidi  \\
Michigan State University\\
\texttt{\{kamalida,kordjams\}@msu.edu} \\
}

\newcommand{\xmark}{\ding{55}} 

\begin{document}

\maketitle

\begin{abstract}
Modern Vision-Language Models (VLMs) have achieved impressive performance in various tasks, yet they often struggle with compositional reasoning, the ability to decompose and recombine concepts to solve novel problems. While neuro-symbolic approaches offer a promising direction, they are typically constrained by crisp logical execution or predefined predicates, which limit flexibility. In this work, we introduce NePTune, a neuro-symbolic framework that overcomes these limitations through a hybrid execution model that integrates the perception capabilities of foundation vision models with the compositional expressiveness of symbolic reasoning. NePTune dynamically translates natural language queries into executable Python programs that blend imperative control flow with soft logic operators capable of reasoning over VLM-generated uncertainty. Operating in a training-free manner, NePTune, with a modular design, decouples perception from reasoning, yet its differentiable operations support fine-tuning. We evaluate NePTune on multiple visual reasoning benchmarks and various domains, utilizing adversarial tests, and demonstrate a significant improvement over strong base models, as well as its effective compositional generalization and adaptation capabilities in novel environments.
\end{abstract}

\begin{figure}[h!]
    \centering
    \includegraphics[width=0.85\linewidth]{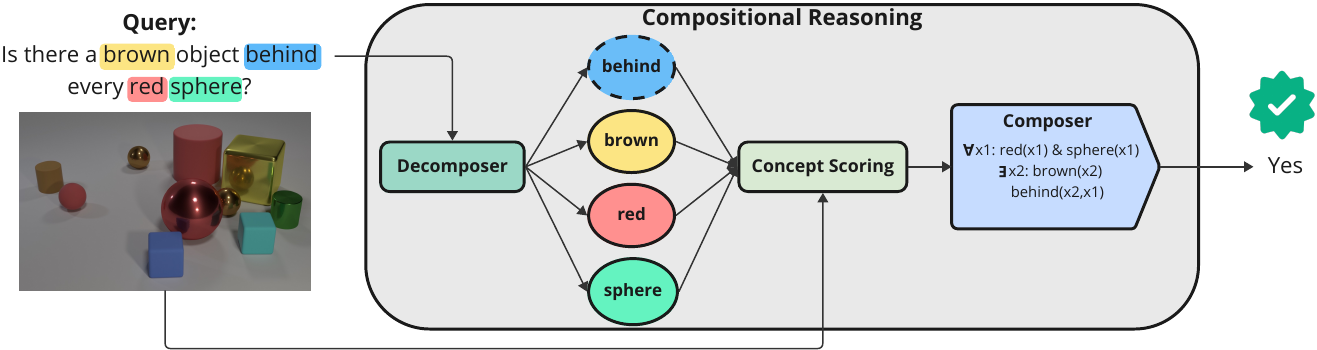}
    \caption{A natural language query 
(``\textit{Is there a brown object behind every red sphere?}'') is \textit{decomposed} into symbolic 
concepts, such as $red$, and $sphere$. 
These concepts are then \textit{composed} to enable explicit reasoning over objects and their 
relations. This illustrates how complex queries can be mapped into structured logical forms.}

    \label{fig:symoblic}
    
\end{figure}

\section{Introduction}

One key aspect of intelligence is the ability to generalize and compose known basic components to solve novel, complex problems. In vision-language reasoning, this capacity for compositional generalization is crucial. Humans easily decompose a complex query like \textit{``Is there a brown object behind every red sphere?"} into its constituent concepts and reason about their relationships~\citep{partee1984compositionality}, as illustrated in Figure~\ref{fig:symoblic}. However, modern vision-language models with transformer-based architectures often fail at dealing with novel compositions, revealing a gap between their pattern-matching skill and robust, human-like understanding~\citep{zhu2022generalization}.

Recent studies have demonstrated the superiority of neuro-symbolic (NeSy) methods over end-to-end neural architectures for reasoning beyond their observed data~\citep{nesycoco,yun2022vision}. In recent years, methods such as VisProg~\citep{Gupta2022VisProg} and ViperGPT~\citep{suris2023vipergpt} have leveraged Large Language Models (LLMs) to produce programs from language queries, breaking down complex questions into a sequence of executable steps. These programs are then run using pre-trained vision models for perceptual grounding. Although powerful, this paradigm has key shortcomings. First, these systems often rely on a sequence of crisp, intermediate decisions, creating brittle pipelines that are sensitive to perceptual errors. A single mistake in an early step (e.g., misidentifying an object) can cause the entire reasoning chain to fail. Second, these pipelines are inference-only and non-differentiable, which prevents them from adapting to new domains where pre-trained models may not perform well. On the other hand, methods such as LEFT~\citep{hsu2024s} and NeSyCoCo~\citep{nesycoco} employ differentiable declarative symbolic reasoning, but their predicates are limited to the set of learned predicates from training on the target domain, which limits their zero-shot applicability. Table~\ref{tab:comparison_compact} summarizes this landscape and highlights where NePTune differs.

\begin{table}[h]
\caption{Comparison of recent neuro-symbolic frameworks. We analyze the reasoning scope (local vs global), types of supported predicates, and predicate source. \textbf{VFM:} vision foundation models, e.g. XVLM; \textbf{VLM:} vision language models, e.g. Qwen2VL.}
\label{tab:comparison_compact}

\centering
\footnotesize
\resizebox{0.9\textwidth}{!}{%
\begin{tabular}{@{}llcccccccc@{}}
\toprule
 & & & \multicolumn{4}{c}{\textbf{Predicate Types}} & \multicolumn{2}{c}{\textbf{Predicates}} \\
 \cmidrule(lr){4-7} \cmidrule(lr){8-9}
\textbf{Model} & \textbf{Reasoning} & \textbf{Supported Predicates} & \textbf{Class} & \textbf{Attr.} & \textbf{Rel.} & \textbf{Spatial} & \textbf{Pretrained} & \textbf{Trainable} \\ \midrule

VisProg & Local & Predefined Modules & \checkmark & \checkmark & \xmark & \checkmark & Modules & \xmark \\
ViperGPT & Local & Dynamic & \checkmark & \checkmark & \xmark & \checkmark & VFMs  & \xmark \\
LEFT & Global & Limited to Training Data & \checkmark & \checkmark & \checkmark & \checkmark & \xmark & \checkmark \\
NeSyCoCo & Global & Limited to Similar Data as Training & \checkmark & \checkmark & \checkmark & \checkmark & \xmark & \checkmark \\
NAVER & Global & Dynamic & \checkmark & \checkmark & \checkmark & \checkmark & VFMs & \xmark \\
\textbf{NePTune} & Hybrid & Dynamic & \checkmark & \checkmark & \checkmark & \checkmark & VLMs & \checkmark \\ \bottomrule
\end{tabular}%
}
\end{table}

To address the challenges of expressive compositional inference and domain adaptation, we propose NePTune, a neuro-symbolic visual reasoning framework. NePTune leverages an LLM to generate expressive Python programs where its execution does not solely rely on crisp decisions. Instead, it performs both soft compositional reasoning on the continuous scores provided by a VLM under uncertainty and imperative sequential reasoning. The result is a framework that decouples reasoning from the perception of atomic concepts, leading to remarkable generalization. 

\begin{calloutbox}[Contributions.]
\textbf{1. A Hybrid Neuro-Symbolic Execution Model:} We propose a novel framework that seamlessly combines the imperative control flow of Python with soft compositional logic, enabling complex reasoning under uncertainty.

\textbf{2. Domain Adaptable Framework:} We present a modular system that uses an LLM to generate programs on the fly, eliminating the need for predefined predicates and enabling zero-shot generalization as well as neuro-symbolic fine-tuning for domain adaptation.

\textbf{3. Strong Compositional Generalization:} We demonstrate through extensive experiments that our approach significantly outperforms existing methods, particularly on challenging domain-shift benchmarks, showing highly robust results for vision-language reasoning.

\end{calloutbox}

\section{Related Work}

NePTune is a framework that integrates multi-modal foundation models and logical reasoning to achieve compositional vision-language reasoning. Therefore, we focus on the three topics below.

\subsection{Compositional Vision-Language Reasoning}
Compositional reasoning is central to building reliable vision–language systems~\citep{sinha2024surveycompositionallearningai, Hupkes2019}, enabling models to represent and execute multi-step, relational structures~\citep{ontanon}. While VLMs achieve strong results across broad tasks ~\citep{xiao2023florence, Qwen2VL, liu2024llavanext}, diagnostic evaluations like CLEVR~\citep{johnson2017clevr} and ReaSCAN~\citep{Wu-ReaSCAN} expose persistent gaps in compositional generalization~\citep{zhu2022generalization} and brittleness under distribution shift~\citep{yun2022vision, li2025multisourcedcompositionalgeneralizationvisual} in end-to-end neural methods. Although improved architectures show promise~\citep{qiu-etal-2021-systematic}, explicit and structured neuro-symbolic methods have been particularly effective on these challenges~\citep{nesycoco}. Motivated by these findings, we adopt a neuro-symbolic approach that performs explicit compositional reasoning to achieve more robust and generalizable visual reasoning.

\subsection{Neuro-Symbolic Approaches}
Neuro-symbolic methods integrate neural perception with symbolic reasoning. Recent approaches utilize LLMs as semantic parsers to interpret linguistic queries. There are two common strategies for such methods. One line of work, including VisProg~\citep{Gupta2022VisProg} and ViperGPT~\citep{suris2023vipergpt}, generates imperative code for a sequence of local reasoning steps that rely on crisp decisions. While powerful, the sequential nature of these methods limits global reasoning under the uncertainty of these local decisions. The second strategy, employed by models such as NSCL~\citep{nscl}, LEFT~\citep{hsu2024s}, and NeSyCoCo~\citep{nesycoco}, focuses on generating a declarative, symbolic representation of the query to enable global reasoning. However, these frameworks rely on training data for concept learning and grounding in each domain, and are therefore constrained by the observed predicates in the training data. While NeSyCoCo improves concept generalization by utilizing a predicate embedding model and covers the lexical variety of natural language, it remains limited to concepts similar to those in its training data. NAVER~\citep{cai2025naver}, on the other hand, emphasizes orchestration through a finite-state controller that manages neural perception and utilizes ProbLog for global reasoning in referring expression grounding. However, our findings indicate that generating ProbLog queries is more error-prone for LLMs compared to Python code. Our aim in this work is to synthesize the advantages of both approaches. NePTune is a neuro-symbolic framework that performs both imperative and soft compositional reasoning over atomic predicates. Unlike works like NeSyCoCo and LEFT, the training is not necessary since the concept scores can be obtained zero-shot by harnessing the power of VLMs. However, with differentiable composition functions, it offers optional fine-tuning for domain adaptability by providing the capacity for both compositional inference as well as neuro-symbolic training.

\subsection{Predicate Level Concept Understanding}

A critical distinction between reasoning frameworks lies in how they derive and utilize concept-level understanding. General-purpose VLMs~\citep{Qwen2VL,liu2024llavanext} learn concepts implicitly through end-to-end training, embedding this knowledge directly into the model's weights. In contrast, neuro-symbolic methods externalize this process. A program generating framework, such as VisProg~\citep{Gupta2022VisProg}, utilizes the concept grounding through a library of specialized, trained, hard-coded vision APIs, which return discrete labels or values. Similarly, ViperGPT~\citep{suris2023vipergpt} utilizes foundation vision-language models, such as XVLM~\citep{xvlm}, to obtain labels and discrete decisions. Another class of methods, including LEFT and NeSyCoCo, trains an MLP to generate continuous concept scores, representing the relations or attributes of an object in the visual regions. Our approach utilizes visual prompting, highlighting parts of the image via bounding boxes, to elicit referential concept scores from VLMs, which then serve as scores within our framework for reasoning under uncertainty.

\section{Methodology}

\begin{figure}[h]
    \centering
    \includegraphics[width=\linewidth]{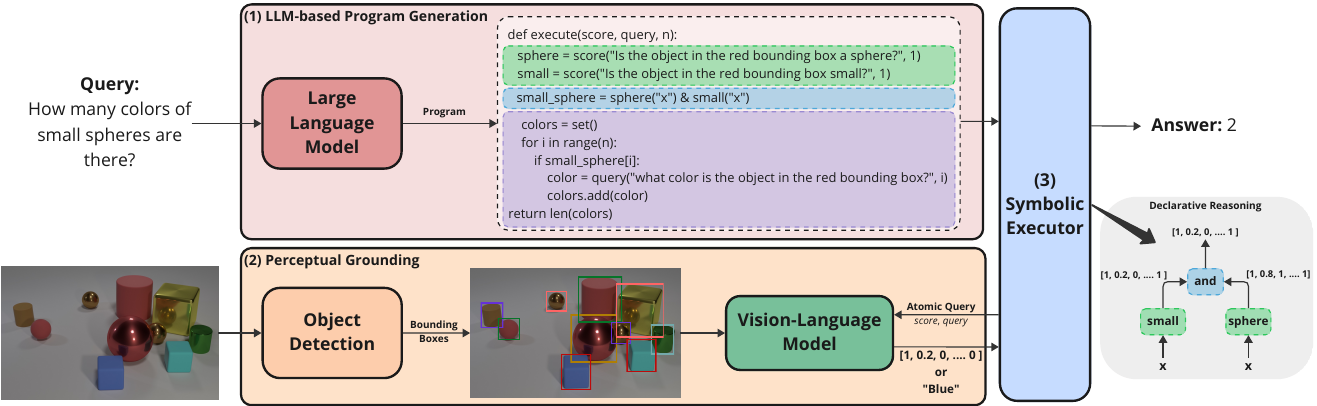}
    \caption{
    NePTune overview. Given an image and a query, the 
\textbf{(1) LLM-based Program Generation} converts the natural language query to a Pythonic program. Then 
\textbf{(2) Perceptual Grounding} extracts the object bounding boxes. The  \textbf{(3) Symbolic Executor} then runs the Python code to reason over \tcbox[
  inlinebox=lightgreen,
  colframe=lightgreen!70!black,   
  boxrule=1pt,              
  arc=2pt,                  
  left=0.5pt,right=0.5pt,       
  top=1pt,bottom=1pt,
  frame style={dashed} 
]{concepts} extracted from the VLM using both \tcbox[
  inlinebox=lightblue,
  colframe=lightblue!70!black,   
  boxrule=1pt,              
  arc=2pt,                  
  left=0.5pt,right=0.5pt,       
  top=1pt,bottom=1pt,
  frame style={dashed} 
]{soft composition} and \tcbox[
  inlinebox=lightpurple,
  colframe=lightpurple!70!black,   
  boxrule=1pt,              
  arc=2pt,                  
  left=0.5pt,right=0.5pt,       
  top=1pt,bottom=1pt,
  frame style={dashed} 
]{imperative logic} to derive the final answer.}
    \label{fig:nesypython}
\end{figure}

The NePTune framework performs visual reasoning via three core components, as shown in Figure \ref{fig:nesypython}. The process begins with the \textbf{LLM-based Program Generator}, which translates a natural language query into both a Python program and a set of relevant object names. These names are passed to the \textbf{Perceptual Grounding} to detect all relevant candidate objects in the scene. Finally, the \textbf{Symbolic Executor} runs the generated program. During execution, it interacts with the Grounding Interface to obtain atomic concept scores and reason over them to get the final answer.

\subsection{Component 1: LLM-based Program Generation}

The first component of NePTune addresses semantic parsing. We leverage an LLM as a powerful few-shot parser to convert a natural language query into a formal, executable Python program. Given a query like ``\textit{Is there a big brown dog?}", the LLM is prompted to decompose it into a multi-step program that first identifies a set of candidate ``dogs", and then reasons over the composition of atomic concepts such as ``big", ``dog", and ``brown" to get the final answer. In the same LLM call, it also extracts object names for the region proposals.  We chose Python as the target symbolic language for two key reasons. First, its Turing-complete nature, including rich control flow structures like loops and conditionals, is essential for expressing complex procedural reasoning. Second, the prevalence of open-source Python code makes LLMs particularly adept at generating it. 

\subsection{Component 2: Perceptual Grounding}
This component connects the symbolic program to the visual world. It consists of two main parts: an object proposal module and a concept grounding module.

\subsubsection{Object Proposal Generation}
To ground objects in the scene to the concepts such as \texttt{blue} in the program, we must first identify all potentially relevant objects in the scene. We use the object names extracted by the LLM. For example, from ``\textit{Are there more cats to the left of the vase or dogs?}", the LLM extracts ``dog", ``vase", and ``cat". We then feed these object names into Grounding DINO~\citep{liu2023grounding}, a zero-shot object detection model that takes the image and the list of object names as input and outputs a set of bounding boxes for all matching objects. We note that while this textual query-based proposal is chosen for computational efficiency in real-world applications, our framework can operate with any reasonable region proposal method.

\subsubsection{Concept Grounding Interface}
Once a query is translated into a program, we must ground its atomic predicates to the visual content of the image, such as the objects in the scene. We handle this \emph{perceptual grounding} using a VLM through two primary simplified interfaces \texttt{\textbf{score}} and \texttt{\textbf{query}}, where the image and bounding boxes are abstracted away with global variables for simplicity. The formal definitions of these grounding functions are as follows:

     \textbf{\texttt{score(query:str, num\_objects:int)}}: This function is the core of our reasoner. It takes a natural language question, an image, and related bounding boxes, and returns the concept probabilities for related object bounding boxes. This score is the VLM's normalized confidence for a "Yes" answer, computed from the logits of the "Yes" and "No" tokens. Formally, given an image $I$, set of $N$ detected objects, a visual prompt $v$, and atomic query $q_a$, the score $s$ is:
    \[ s = p(\text{"Yes"} | I, v, q_a) = \frac{e^{\text{logit}(\text{"Yes"})}}{e^{\text{logit}(\text{"Yes"})} + e^{\text{logit}(\text{"No"})}} \]
We treat these outputs as probability scores. To generate an object-centric concept score or answer via a VLM, we represent a symbolic predicate with a natural language question. For example, for the symbolic predicate \texttt{blue}, we use ``\textit{Is the object in the red bounding box blue?}". The bounding box $b$ provided to the functions will be drawn on target objects as a \emph{visual prompt}, and it is controlled by the \texttt{num\_objects} parameter, which specifies the predicate type:
\begin{itemize}
    \item \textbf{\texttt{num\_objects=0}}: For queries about the entire image (e.g., ``\textit{Is the photo taken indoors?}"). No visual prompt is used, and the output is a scalar probability.
    
    \item \textbf{\texttt{num\_objects=1}}: For single-object queries, we mark the target object with a red bounding box around it as a visual prompt (e.g., ``\textit{Is the object in the red bounding box brown?}"). The output is a vector of size $N$.
    
    \item \textbf{\texttt{num\_objects=2}}: For multi-object (relational) queries, we mark the main object with a red bounding box, and the secondary object with a green bounding box, as the visual prompt (e.g., ``\textit{Is the person in the red box talking to the person in the green box?}"). The output is an $N \times N$ matrix.
\end{itemize}

\textbf{\texttt{query(query:str, object\_id:int)}}: 
    This query function is used for tasks requiring open-ended answers. It takes a natural language question, an image, and an optional visual prompt, and returns a natural language string. Formally, given an image $I$, visual prompt $v$, and atomic query $q_a$, the generated answer $A$:
    \[ A = \operatorname*{argmax}_{\text{response}} p(\text{response} | I, v, q_a) \]
    
The answer $A$ serves two purposes. First, it can provide the final output for questions that ask "\textit{what}" or "\textit{which}" (e.g., `return query("\textit{What color is the object in the red bounding box?}")`). Second, it can be used as an intermediate variable to enable complex conditional constructs within the generated Python program, allowing the reasoning path to change based on the VLM's perception (e.g., `if query("\textit{What shape is the object in the red bounding box?}") == "\textit{cube}": ...`).

\subsection{Component 3: Symbolic Executor}

\label{sec:component3}

The final component of NePTune is the \textbf{Symbolic Executor}, which runs the LLM-generated program. A key innovation of our framework is its hybrid execution model, which integrates two distinct reasoning modes.

\noindent \textbf{1) Soft Compositional Reasoning:} To reason about visual concepts themselves, we employ a set of soft logical operations based on fuzzy logic principles. Instead of operating on binary true/false values, these operations work directly on the uncertainty scores obtained from the VLM. This is implemented through our custom data structures, which encapsulate the scores for a given predicate and overload standard Python operators (\texttt{\&} for AND, \texttt{|} for OR) to perform the corresponding logical operations. Details of these operations are shown in Table \ref{tab:expressions}. For example, when the program executes \texttt{brown \& dog}, our framework takes the element-wise minimum of the scores in the two corresponding tensors, implementing the fuzzy t-norm for conjunction.

\begin{table}[h!]
\caption{Mathematical Expressions: Logical Forms, Descriptions, and Differentiable Implementations. Here $\alpha$ represents an object-centric or scalar probabilistic score, $\beta$ represents a relation probability score, $\tau=0.25$ is a temperature parameter, and $\gamma=0.25$ is a margin.}
\label{tab:expressions}
\centering
\resizebox{0.8\textwidth}{!}{%
\begin{tabular}{@{}lllc@{}}
\toprule
\textbf{Syntax} & \textbf{Logical Form} & \textbf{Description} & \textbf{Differentiable Implementation} \\
\midrule
$\alpha_x.\texttt{exists}()$ & $\exists x \, \alpha_x$ & Existential quantification & $\max(\alpha_x)$ \\
$\alpha_x.\texttt{forall}()$ & $\forall x \, \alpha_x$ & Universal quantification & $\min(\alpha_x)$ \\
$\alpha_x \,\&\, \alpha_y$ & $\alpha_x \land \alpha_y$ & Logical conjunction & $\min(\alpha_x, \alpha_y)$ \\
$\alpha_x \,\&\, \beta_{xy}$ & $\alpha_x \land \beta_{xy}$ & Relational conjunction & $\sum_{y} \alpha_x \cdot \beta_{xy}$ \\
$\alpha_x \,|\, \alpha_y$ & $\alpha_x \lor \alpha_y$ & Logical disjunction & $\max(\alpha_x, \alpha_y)$ \\
$\alpha_x.\texttt{implies}(\alpha_y)$ & $\alpha_x \rightarrow \alpha_y$ & Logical implication & $\max(1-\alpha_x, \alpha_y)$ \\
$\texttt{not } \alpha_x$ & $\neg \alpha_x$ & Logical negation & $1 - \alpha_x$ \\
$\alpha_x.\texttt{iota}(\text{var})$ & $\iota(\text{var}, \alpha_x)$ & Best match & $\operatorname{softmax}(\alpha_x)$ \\
$\alpha_x.\texttt{count}()$ & $\text{count}(\alpha_x)$ & Counting elements & $\sum \alpha_x$ \\
$s_1 == s_2$ & $s_1 = s_2$ & Scalar equality & $\sigma\left( \frac{\tau(\gamma - |s_1 - s_2|)}{\gamma} \right)$ \\
$s_1 > s_2$ & $s_1 > s_2$ & Scalar inequality & $\sigma\left( \tau(s_1 - s_2 - 1 + \gamma) \right)$ \\
\bottomrule
\end{tabular}
}

\end{table}

 \noindent \textbf{2) Imperative Reasoning:} Our symbolic executor leverages a standard Python interpreter to handle the program's overall structure and control flow. This is possible since we have defined iteration and Boolean operations on the concept objects, allowing for complex, procedural logic, including conditionals (\emph{if}/\emph{else}), loops (\emph{for}), and variable assignments, giving our framework the full expressive power of a general-purpose programming language. While some of these operations such as counting over string sets (as shown in Figure~\ref{fig:nesypython}) can break the computation graph. This hybrid design enables the system to reason fluidly under uncertainty while fully leveraging the expressive power of a general-purpose programming language.

\section{Experiments}

We structure our experiments around four research questions. \textbf{RQ1:} How does NePTune perform on zero-shot compositional reasoning on synthetic data? (Experiment 1), \textbf{RQ2:} How does NePTune perform on complex human-generated questions? (Experiment 2), \textbf{RQ3:} How does {NePTune} perform in grounding referring expressions in natural images? (Experiment 3), and \textbf{RQ4:} How does NePTune perform under domain shift and adapt to an unseen environment? (Experiment 4).
To address these questions, we evaluate NePTune across a diverse set of datasets covering synthetic, human-annotated, and realistic environments, spanning both question answering and referring expression grounding tasks. Details of the datasets and tasks are provided in Appendix~\ref{app:datasets}.
\subsection{Experiment 1: Core Compositional Reasoning}
\label{exp:1}
\begin{table}[h!]
\centering
\caption{Accuracy comparison across CLEVR question categories. Results are shown by reasoning type and model paradigm, including zero-shot, end-to-end, and neuro-symbolic approaches.}

\resizebox{0.75\textwidth}{!}{%
\begin{tabular}{lccc|cc}
\toprule
\textbf{} & \textbf{InternVL2.5} & \textbf{NePTune} & \textbf{ViperGPT} & \textbf{NeSyCoCo} & \textbf{LEFT} \\
\midrule
\textbf{Training} & Zero-Shot & Zero-Shot & Zero-Shot & Trained & Trained \\
\textbf{Category} & End-to-End & NeSy & NeSy & NeSy & NeSy \\
\midrule
Final Accuracy & 90.25 & \textbf{92.65} (\textcolor{green}{$\uparrow$ 2.40}) & 36.05 & 99.68 & 99.50 \\
\midrule
Exist & 87.10 & \textbf{93.19} (\textcolor{green}{$\uparrow$ 6.09}) & 48.75 & 99.28 & 98.92 \\
Query Attribute & \textbf{98.26} & 96.81 (\textcolor{red}{$\downarrow$ 1.45}) & 29.42 & 100.00 & 99.86 \\
Compare Attribute & \textbf{98.61} & 91.94 (\textcolor{red}{$\downarrow$ 6.67}) & 53.06 & 99.44 & 99.72 \\
Count & 74.60 & \textbf{87.10} (\textcolor{green}{$\uparrow$ 12.50}) & 21.37 & 99.79 & 98.99 \\
Compare Number & 90.86 & \textbf{92.57} (\textcolor{green}{$\uparrow$ 1.71}) & 48.57 & 100.00 & 100.00 \\
\bottomrule
\end{tabular}
}
\label{tab:clevr}
\end{table}

To evaluate NePTune's reasoning capabilities, we evaluate its performance on the \textbf{CLEVR} benchmark. CLEVR is a standard benchmark for compositional reasoning that features synthetic 3D-rendered images and questions testing compositional visual reasoning. Details of the question categories are provided in Appendix~\ref{app:datasets}. The results in Table \ref{tab:clevr} show that, within the family of neuro-symbolic systems, NePTune establishes itself as the strongest zero-shot method, achieving 92.65\% accuracy compared to only 36.05\% for ViperGPT. While trained approaches such as NeSyCoCo and LEFT nearly saturate CLEVR (99\%), NePTune demonstrates that competitive compositional reasoning can be achieved without dataset-specific supervision. Compared to its backbone VLM, InternVL2.5, NePTune still yields improvements raising overall accuracy from 90.25\% to 92.65\% (\textcolor{green}{$\uparrow$ 2.40\%}). The largest gains appear in quantitative categories where explicit compositional structure is most useful: \textit{Count} rises from 74.60\% to 87.10\% (\textcolor{green}{$\uparrow$ 12.50\%}), and \textit{Compare Number} increases from 90.86\% to 92.57\% (\textcolor{green}{$\uparrow$ 1.71\%}). We also see a notable improvement on \textit{Exist} from 87.10\% to 93.19\% (\textcolor{green}{$\uparrow$ 6.09\%}), consistent with the executor reducing spurious correlation when filtering by attributes and relations. In contrast, attribute-heavy categories regress: \textit{Query Attribute} drops by \textcolor{red}{$\downarrow$ 1.45\%} points and \textit{Compare Attribute} drops by \textcolor{red}{$\downarrow$ 6.67\%} points. A closer look at concept-level accuracy revealed that analogical concepts such as \textit{same color} or \textit{same shape} remain among the most challenging to capture in this benchmark. More discussion of atomic concept evaluation is provided in Section~\ref{sec:analysis}.

\begin{table}[h]
\centering
{

\caption{Accuracy on CLEVR extension tasks. Methods marked with $^{\dagger}$ use ground-truth programs. Improvements from the backbone VLM (InternVL2.5) are marked with the arrow sign (\textcolor{green}{$\uparrow$}).}
\label{tab:clevr-extensions}

  \resizebox{0.6\textwidth}{!}{%
\begin{tabular}{l|lccc}
\toprule
& \textbf{Method } & \textbf{Ref(\%)} &  \textbf{Puzzles(\%)} &  \textbf{RPM(\%)} \\ 
\toprule
\multirow{3}{*}{\rotatebox{90}{Trained}} 
  & NeSyCoCo$^{\dagger}$ & \textbf{100.00} & \textbf{95.00} & \textbf{100.00} \\ 
  & NeSyCoCo & 94.00 & 94.00 & 74.00 \\ 
  & LEFT & 94.00 & 85.00 & 87.00 \\ 
\midrule
\multirow{7}{*}{\rotatebox{90}{Zero-shot}} 
  & Qwen2VL.5-8B & 21.00 & 43.00 & \textbf{53.00} \\
  & InternVL2.5-8B & \textbf{27.00} & \textbf{52.00} & 47.00 \\
  & Ovis1.6-9B & 4.00 & 47.00 & 49.00 \\
  \cmidrule(lr){2-5}
  & ViperGPT  & 8.00 & 34.00 & 4.00 \\ 
  & VisProg & 35.00 & 27.00 & 51.00 \\ 
  
  & NePTune$^{\dagger}$& \textbf{99.00} (\textcolor{green}{$\uparrow$ 72}) & \textbf{65.00}  (\textcolor{green}{$\uparrow$ 13})  & \textbf{99.00}  (\textcolor{green}{$\uparrow$ 52})  \\ 
  & NePTune & 91.00 (\textcolor{green}{$\uparrow$ 64}) & 60.00 (\textcolor{green}{$\uparrow$ 8}) & 80.00 (\textcolor{green}{$\uparrow$ 33}) \\ 
\bottomrule
\end{tabular}
}
}

\end{table}
\vspace{-2mm}
In addition to the original CLEVR benchmark, we evaluate our method on various compositional challenges based on the CLEVR environment introduced in LEFT \citep{hsu2024s}, including referring expressions \textbf{CLEVR-Ref}, visual puzzles \textbf{CLEVR-Puzzles}, and Raven's Progressive Matrices \textbf{CLEVR-RPM}. As shown in Table \ref{tab:clevr-extensions}, NePTune shows the highest performance among the zero-shot methods and outperforms the end-to-end VLMs. In addition, compared to NeSyCoCo and LEFT with trained concepts, we show competitive performance.

\subsection{Experiment 2: Complex Human Queries}
\label{exp:2}
\begin{wraptable}{r}{0.35\textwidth}  
  
  \vspace{-4.5mm}                       
\caption{Accuracy on the CLEVR-Humans (CH) dataset.}
  \label{tab:clevr-humans}

  \centering
  \resizebox{0.29\textwidth}{!}{%
    \begin{tabular}{l|lc}
      \toprule
      & \textbf{Method}  & \textbf{CH (\%)} \\
      \toprule
      \multirow{3}{*}{\rotatebox[origin=c]{90}{\parbox{1cm}{\centering Trained}}} 
        & LEFT & 56.69 \\
        & NeSyCoCo & 56.12 \\
        & MDETR & \textbf{81.73} \\
      \midrule
      \multirow{5}{*}{\rotatebox[origin=c]{90}{\parbox{1.8cm}{\centering Zero-shot}}} 
        & Qwen2VL-7B & 84.12 \\
        & InternVL2.5-8B  & \textbf{85.95} \\
        & Ovis1.6-9B  & 79.96 \\
      \cmidrule(lr){2-3}
        & ViperGPT & 31.05 \\
        & NePTune & \textbf{87.67} \\
      \bottomrule
    \end{tabular}
  }

\end{wraptable}

Here, we evaluate NePTune in a more complex and diverse human-generated language. We utilize the \textbf{CLEVR-Humans}~\citep{johnson2017inferringexecutingprogramsvisual} benchmark to evaluate and compare NePTune against other neuro-symbolic and end-to-end models. As shown in Table \ref{tab:clevr-humans}, NePTune significantly outperforms prior neuro-symbolic methods such as LEFT and NeSyCoCo by a large margin of \textcolor{green}{$\uparrow$~30.98\%} even compared to end-to-end methods such as MDETR~\citep{kamath2021mdetr}. Furthermore, it improves upon its powerful end-to-end backbone (InternVL2.5-8B) by \textcolor{green}{$\uparrow$~1.72\%}, demonstrating its effectiveness on complex, human-generated questions.

\subsection{Experiment 3: Real-world Images}
\label{exp:3}

\begin{wraptable}{r}{0.3\textwidth}  
  \vspace{-15pt}   
  \caption{Accuracy on Ref-Adv dataset.}
  \centering  
  \setlength{\tabcolsep}{1mm}          
  \resizebox{0.3\textwidth}{!}{%
    \begin{tabular}{@{}lc@{}}
      \toprule
      \textbf{Method} & \textbf{Ref-Adv (\%)} \\
      \toprule
      Grounding DINO-B & 60.85 \\
      Florence2-L & 71.73 \\
      Ovis1.6-9B & 30.70 \\
      InternVL2-8B & 72.92 \\
      InternVL2.5-8B & 76.13 \\
      \cmidrule(lr){1-2}
      ViperGPT & 60.66 \\
      NAVER$^{\dagger}$ & 36.45 \\
      NAVER & 65.13 \\
      NePTune$^{\ddagger}$ & 63.71 \\
      \hspace{1mm} + Verification & 75.54 \\
      NePTune & 71.57 \\
      \hspace{1mm} + Verification & \textbf{78.08} \\
      NePTune (1B) & 60.69 \\
      \hspace{1mm} + Fine-tuning & 68.06 $\pm$ 0.56 \\
      \hspace{1mm} + Verification & 74.59 $\pm$ 0.12  \\
      \bottomrule
    \end{tabular}  
    }
  \label{tab:ref-adv}
  \vspace{-5mm}                       
\end{wraptable}

While various versions of CLEVR used in the aforementioned experiments are strong testbeds for compositional reasoning, after all, these are toy environments with limited diversity. To demonstrate NePTune's ability to operate in realistic environments, we evaluate it on Referring Expression Grounding (REG) of real-world images using the \textbf{RefCOCO-Adversarial}~\citep{refadv} (Ref-Adv) benchmark. On Ref-Adv, NePTune is the strongest zero-shot reasoner among symbolic baselines such as ViperGPT and NAVER. To ensure a fair comparison with NAVER, we follow its setup and use the same backbones (Grounding DINO + XVLM + InternVL2-8B), denoted as NePTune$^{\ddagger}$. Under this setting, NePTune$^{\ddagger}$ reaches 63.71 vs.\ 36.45 for NAVER$^{\dagger}$ (execution only), and with a simple verification (similar to NAVER) attains 71.57 vs.\ 65.13 for NAVER.  Compared to end-to-end methods, NePTune surpasses all the specialized grounding methods such as Grounding DINO and Florence2~\citep{xiao2023florence2}. Additionally, compared to their VLM backbones, both NePTune and NePTune$^{\ddagger}$ with verification surpass their respective backbones. Details on verification are available in Appendix~\ref{app:verification}.

\subsection{Experiment 4: Generalization and Adaptation}
\label{exp:4}
When using popular benchmarks, as we did in Experiments 1-3, the possible data contamination makes zero-shot performance less reliable. In this experiment, we test NePTune's ability to generalize and adapt to novel environments using a less popular photo-realistic gaming environment that is \textbf{Ref-GTA}~\citep{refgta} benchmark. Ref-GTA is a challenging REG benchmark with images from a game simulation, which creates a significant domain shift from the natural images on which most VLMs are pre-trained. \begin{wraptable}{r}{0.3\textwidth}  
\vspace{-2mm}
    \caption{Accuracy on Ref-GTA benchmark.}
  \centering
  \resizebox{0.3\textwidth}{!}{%
    \begin{tabular}{@{}lc@{}}
      \toprule
      \textbf{Method} & \textbf{Ref-GTA(\%)} \\
      \midrule
      GroundingDINO-B & 27.90 \\
      Florence2-L & 58.65 \\
      Ovis1.6-9B & 2.78 \\
      InternVL2.5-8B & 6.95 \\
      InternVL2.5-1B & 1.64 \\
      \hspace{1mm} + Fine-tuning & 32.61 $\pm$ 0.35 \\
      \cmidrule(lr){1-2}
      ViperGPT & 1.40 \\
      NAVER$^{\dagger}$ & 54.84 \\
      NAVER & 58.73 \\
      NePTune$^{\ddagger}$ & 62.73 \\
      NePTune & \textbf{69.69} \\
      NePTune (1B) & 34.92 \\
      \hspace{1mm} + Fine-tuning & 69.90 $\pm$ 1.16 \\
      \bottomrule
    \end{tabular}
}

  \label{tab:ref-gta}
\vspace{-2mm}
\end{wraptable}Based on the reports of Ovis1.6~\citep{lu2024ovis} and InternVL2.5~\citep{internvl25}, this source is not included in their training data. As shown in Table~\ref{tab:ref-gta}, this domain shift causes a significant performance drop for most methods. The powerful end-to-end InternVL2.5-8B model, for example, fails catastrophically, with its accuracy dropping to 6.95\%. In contrast, NePTune, when paired with the same VLM, achieves a remarkable 69.69\% accuracy. This demonstrates that by utilizing our global symbolic reasoner and atomic-level concept understanding, our framework achieves robustness and compositional generalization that monolithic models lack. Furthermore, while NePTune demonstrates strong zero-shot robustness, its soft operations also enable fine-tuning. As shown in Table~\ref{tab:ref-gta}, by fine-tuning a smaller VLM (InternVL2.5-1B) using our neuro-symbolic differentiable computations with only 1000 samples, we can further improve its performance and obtain 69\% accuracy.
While fine-tuning VLM using original neural training, it only reaches 32\% accuracy. These findings are promising for future research and using neuro-symbolic reasoning as a source of supervision for larger-scale VLM training.  More details on fine-tuning are presented in Appendix~\ref{app:hyperparams}.

\section{Discussion}
\label{sec:analysis}
\paragraph{VLM for Concept Understanding.} 
A central idea of NePTune is to employ VLMs as underlying perception modules and concept grounders. The basic assumption here is that while VLMs are prone to fail at reasoning over complex compositions, they should perform better in perceiving basic concepts. However, basic perception questions require isolating parts of the visual input, a process known as visual prompting, which has been shown to be challenging in complex questions~\citep{cai2024vipllava}. In this section, we aim to evaluate the hypothesis that modern VLMs guided by visual prompts are effective as concept grounders for our purpose. We utilize ground-truth scene graphs from the \textbf{CLEVR}~\citep{johnson2017clevr} and \textbf{Visual Genome}~\citep{krishna2016visualgenomeconnectinglanguage} datasets to automatically generate a set of atomic templated questions. For CLEVR, we create a benchmark using scene graphs of 200 sampled scenes, and for Visual Genome, we randomly sample 1000 questions from 200 scene graphs. We generate simple Yes or No questions about their class, attributes, and relations, such as, ``\textit{Is the object in the bounding box a bird?}".  We then use our \texttt{score} function that employs the underlying VLM to answer these atomic questions. 

\begin{table}[h!]

\centering
\caption{Backbone VLMs performance on CLEVR and VG scene graphs (Micro F1-Score x 100)}
\label{tab:atomic_merged}
  \resizebox{0.65\linewidth}{!}{%

\centering
\small
\begin{tabular}{@{}l*{3}{cc}@{}}
\toprule
\textbf{Category}
& \multicolumn{2}{c}{\textbf{InternVL2.5-8B}}
& \multicolumn{2}{c}{\textbf{Ovis1.6-9B}}
& \multicolumn{2}{c}{\textbf{Qwen2VL-7B}} \\
\cmidrule(lr){2-3} \cmidrule(lr){4-5} \cmidrule(lr){6-7}
& \textbf{CLEVR} & \textbf{VG}
& \textbf{CLEVR} & \textbf{VG}
& \textbf{CLEVR} & \textbf{VG} \\
\midrule
Attribute     & \textbf{97.25} & \textbf{89.37} & 95.87 & 86.75 & 89.27 & 83.45 \\
Class/Object  & 90.00 & 81.76 & \textbf{91.98} & 80.42 & 82.21 & \textbf{83.28} \\
Relation      & \textbf{90.94} & 82.35 & 87.91 & 80.00 & 80.04 & \textbf{83.02} \\
Spatial       & \textbf{89.82} & 93.42 & 86.83 & 93.39 & 89.53 & \textbf{94.41} \\
\cmidrule(lr){1-7}
\textbf{Overall} & \textbf{94.54} & 90.41 & 90.35 & 88.19 & 86.59 & \textbf{90.77} \\
\bottomrule
\end{tabular}
}
\end{table}
\vspace{-2mm}

The results are shown in Table \ref{tab:atomic_merged}. Our analysis reveals that VLMs are particularly strong at identifying object, classes, properties, and spatial relations. However, the performance is weaker on more complex relational concepts, especially for analogical comparisons such as \textit{same size}. This problem is more severe in the Qwen2VL model. 
Although visual prompting can be problematic for general VQA~\citep{cai2024vipllava}, our results show this is not as challenging in our setting for grounding atomic queries. These results demonstrate that VLMs perform significantly better in answering atomic queries compared to complex multi-step queries, as reported in Tables~\ref{tab:clevr-extensions} and \ref{tab:clevr-humans}. For example,  there is a performance gap of up to 67\%  when comparing CLEVR-Ref (27\%) compared to CLEVR average atomic evaluation (94\%) using InternVL2.5. Detailed concept-level results of the experiment are presented in the Appendix~\ref{app:atomic}. Furthermore, the results in Table~\ref{tab:end-to-end} demonstrate the end-to-end performance of different VLMs in NePTune, where backbones show an average improvement of up to \textcolor{green}{$\uparrow$~29\%}.

\begin{table}[h!]
\centering
\caption{Performance of different VLM backbones (end-to-end) and with NePTune across benchmarks. $\Delta$ values indicate gains or losses relative to the raw VLMs.}
\label{tab:end-to-end}
\resizebox{\linewidth}{!}{
\begin{tabular}{@{}lccccccc@{}}
\toprule
\textbf{Model} & \textbf{CLEVR (\%)} & \textbf{Ref-Adv (\%)} & \textbf{Ref-GTA (\%)} & \textbf{Ref (\%)} & \textbf{Puzzles (\%)} & \textbf{RPM (\%)} & \textbf{Avg. (\%)} \\
\midrule
InternVL2.5-8B & 90.25 & 76.13 & 6.95 & 27.00 & 52.00 & 47.00 & 49.89 \\
\hspace{2mm} + NePTune & 92.65 \textcolor{green}{$\uparrow$ 2.40} & 78.08 \textcolor{green}{$\uparrow$ 1.95} & 69.69 \textcolor{green}{$\uparrow$ 62.74} & 91.00 \textcolor{green}{$\uparrow$ 64.00} & 60.00 \textcolor{green}{$\uparrow$ 8.00} & 80.00 \textcolor{green}{$\uparrow$ 33.00} & \textcolor{green}{78.57 ($\uparrow$ 28.68)} \\
\midrule
Ovis1.6-9B & 85.00 & 58.47 & 2.78 & 4.00 & 47.00 & 49.00 & 41.04 \\
\hspace{2mm} + NePTune & 88.80 \textcolor{green}{$\uparrow$ 3.80} & 63.25 \textcolor{green}{$\uparrow$ 4.78} & 56.32 \textcolor{green}{$\uparrow$ 53.54} & 75.00 \textcolor{green}{$\uparrow$ 71.00} & 57.00 \textcolor{green}{$\uparrow$ 10.00} & 80.00 \textcolor{green}{$\uparrow$ 31.00} & \textcolor{green}{70.06 ($\uparrow$ 29.02)} \\
\midrule
Qwen2VL-7B & 93.55 & 81.07 & 1.75 & 21.00 & 43.00 & 53.00 & 48.90 \\
\hspace{2mm} + NePTune & 82.55 \textcolor{red}{$\downarrow$ 11.00} & 80.93 \textcolor{red}{$\downarrow$ 0.14} & 68.87 \textcolor{green}{$\uparrow$ 67.12} & 71.00 \textcolor{green}{$\uparrow$ 50.00} & 53.00 \textcolor{green}{$\uparrow$ 10.00} & 71.00 \textcolor{green}{$\uparrow$ 18.00} & \textcolor{green}{71.23 ($\uparrow$ 22.33)} \\ 
\bottomrule
\end{tabular}
}

\end{table}

\begin{wraptable}{r}{0.35\textwidth}  
\vspace{-4.5mm}
\caption{Ablation study on CLEVR-Humans (CH).}
\label{tab:clevr-humans-ablation}
\centering

\resizebox{0.35\textwidth}{!}{%
\begin{tabular}{@{}lc@{}}
\toprule
\textbf{Ablation Setting}  & \textbf{CH (\%)} \\
\midrule
Declarative + Trained Concepts & 56.12 \\
+ VLM Concepts                 & 68.48 (\textcolor{green}{$\uparrow$ 12.36}) \\
+ Imperative Reasoning          & \textbf{87.67} (\textcolor{green}{$\uparrow$ 19.19}) \\
\bottomrule
\end{tabular}
}

\vspace{-2mm}

\end{wraptable}

\paragraph{Ablation Study.} Table~\ref{tab:clevr-humans-ablation} presents an ablation study of NePTune on CLEVR-Humans, starting from a declarative reasoning backbone. Incorporating InternVL2.5 as the concept scoring module yields a \textcolor{green}{$\uparrow$~12.36} improvement, showing that strong VLMs provide more robust and generalizable concept grounders than trained concept grounders on CLEVR. Building on this, adding NePTune's imperative reasoning further boosts accuracy by \textcolor{green}{$\uparrow$~19.19}, closing the gap left by purely declarative systems.

\paragraph{Choice of Symbolic Program.} We analyze program execution success rates (no syntax or runtime errors) on 500 Ref-Adv and CLEVR-Humans using the GPT-4o~\citep{openai2024gpt4ocard} LLM. As shown in Table \ref{tab:clevr-ext-results}, frameworks that rely on specialized, non-Pythonic syntax, such as LEFT, NeSyCoCo, and NAVER (which uses ProbLog), frequently suffer from generation failures. 
\begin{wraptable}{r}{0.4\textwidth}  
  \vspace{-8pt}                      
\caption{Execution success rate (\%) of different neuro-symbolic methods on Ref-Adv and CLEVR-Humans. Incompatible models are excluded.}
\label{tab:clevr-ext-results}
  \centering

  \resizebox{0.37\textwidth}{!}{

  \begin{tabular}{lcc}
    \toprule
    \textbf{Method} & \textbf{Ref-Adv} & \textbf{CLEVR} \\
    \toprule
    NeSyCoCo    & N/A               & 70.33$\pm$3.47 \\
    LEFT        & N/A               & 64.33$\pm$3.94 \\
    ViperGPT    & 48.67$\pm$1.70    & 95.42$\pm$0.30 \\
    NAVER       & 23.02$\pm$4.71    & N/A \\
    NePTune     & \textbf{98.66}$\pm$0.82 & \textbf{97.24}$\pm$0.85 \\
    \bottomrule
  \end{tabular}
}
  \vspace{-3mm}
\end{wraptable}
We found that the LLM often struggles to correctly translate natural language into their specific, strict formalisms. In contrast, methods that utilize Python, such as ViperGPT and NePTune, demonstrate significantly higher rates of successful program execution. 
However, our analysis of ViperGPT's failures reveals that its purely imperative approach, which often relies on selecting objects by a fixed index, is a primary source of error. The NePTune hybrid reasoner proves to be the most robust. Our analysis shows that the main source of error for NePTune shifts from low-level syntax errors to higher-level logical errors, where the LLM fails to compose the correct sequence of predicates. Qualitative examples are shown in Appendix~\ref{app:qualitative}.

\section{Conclusion}
In this work, we introduce NePTune, a novel neuro-symbolic framework for compositional vision-language reasoning. Our approach leverages a novel hybrid reasoner that combines the imperative control flow of Python with a declarative, probabilistic soft logic that operates directly on the uncertain outputs of a VLM. This design enables our framework to be both highly expressive and robust in the face of perceptual uncertainty.
Our extensive experiments demonstrate the effectiveness of this approach. NePTune demonstrates clear improvement over strong baselines, as well as the underlying VLMs, on complex compositional reasoning benchmarks, in visual question answering, and referring expression, exhibiting remarkable generalization capabilities. While the performance of NePTune is dependent on the quality of its underlying components, such as the LLM for program generation and the VLM for concept grounding, our work demonstrates a flexible and powerful paradigm for building more robust and generalizable AI systems, leveraging these models to elevate their capabilities to a comparatively higher level of performance. Future work could explore methods for more efficient grounding and the integration of our work into visual reasoning orchestration.

\section*{Acknowledgments}

This project is supported by the Office of Naval Research (ONR) grant N00014-23-1-2417. Any opinions, findings, and conclusions or recommendations expressed in this material are those of the authors and do not necessarily reflect the views of Office of Naval Research.

\section*{Ethics Statement}
This work complies with the ICLR Code of Ethics. NePTune builds on publicly available datasets and open-source VLMs, without collecting new human-subject data. Large Language Models were used solely for grammar polishing and minor text improvements.

\section*{Reproducibility Statement}
We detail model design, training, and evaluation in the paper and appendix, including hyperparameters and dataset preprocessing. Ablations and failure analyses clarify component contributions. Anonymized code and scripts will be released to enable reproduction of all reported results.

\bibliography{iclr2026_conference}
\bibliographystyle{iclr2026_conference}

\appendix
\section{Qualitative Analysis}
\label{app:qualitative}

\subsection{Referring Expressions Grounding}
\begin{figure}[h!]
    \centering
    \includegraphics[width=1\linewidth]{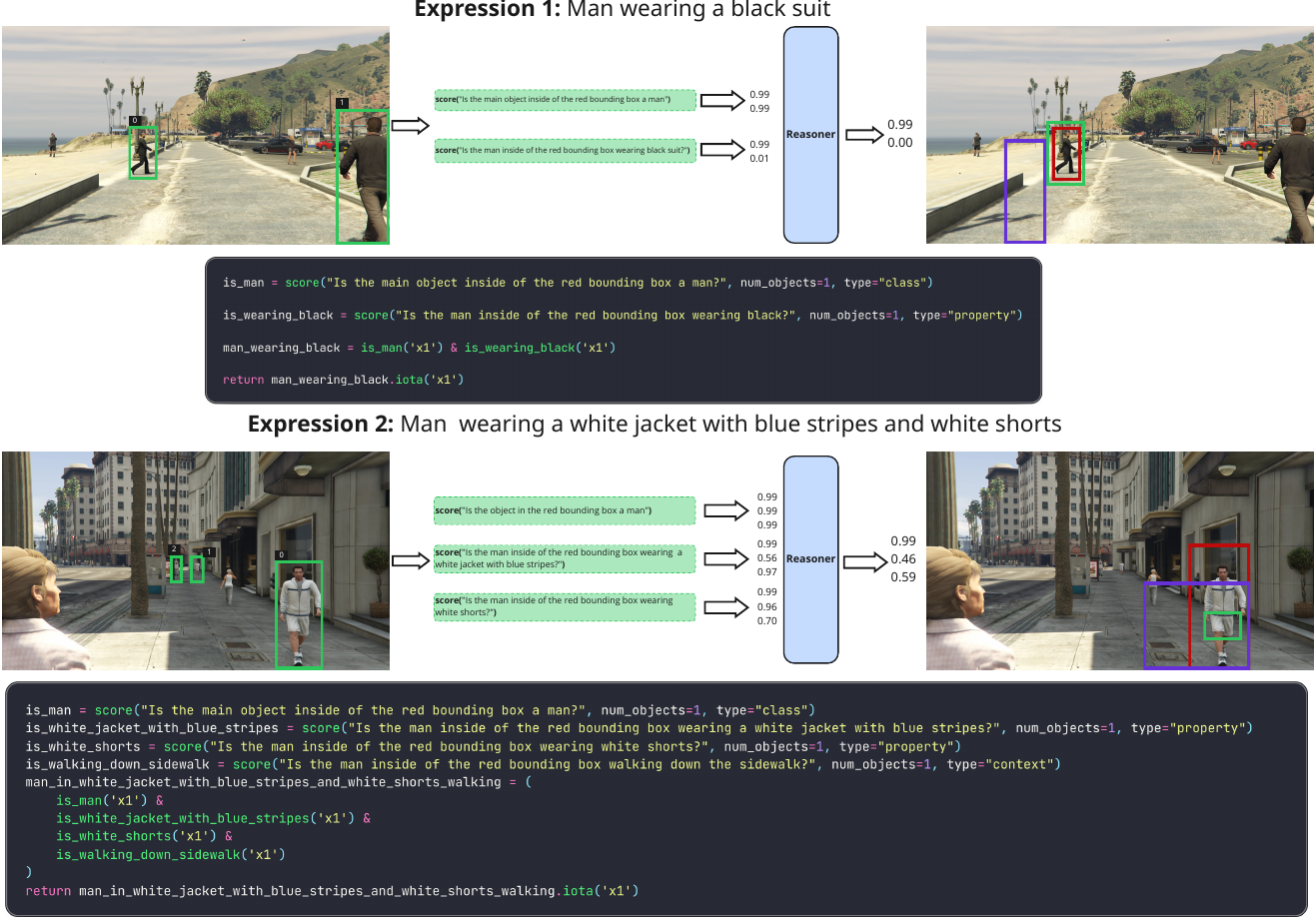}
    \caption{Qualitative examples of NePTune on the RefGTA dataset. Green boxes indicate objects detected by Grounding DINO, blue boxes show objects selected by the VLM (InternVL2.5-8B), and red boxes highlight the final selections made by NePTune.}
    \label{fig:qualitative-refgta}
    
\end{figure}

Examples of NePTune on Ref-GTA are shown in Figure~\ref{fig:qualitative-refgta}. \textbf{Expressions 1 and 2} show the superiority of compositional reasoning compared to the VLM and object detection backbones. They demonstrate how the VLM fails to correctly locate the object in the new environments, while atomic concepts are correctly scored by the VLM.

\subsection{Visual Questions Answering}
\begin{figure}[h!]
    \centering
    
    \includegraphics[width=1\linewidth]{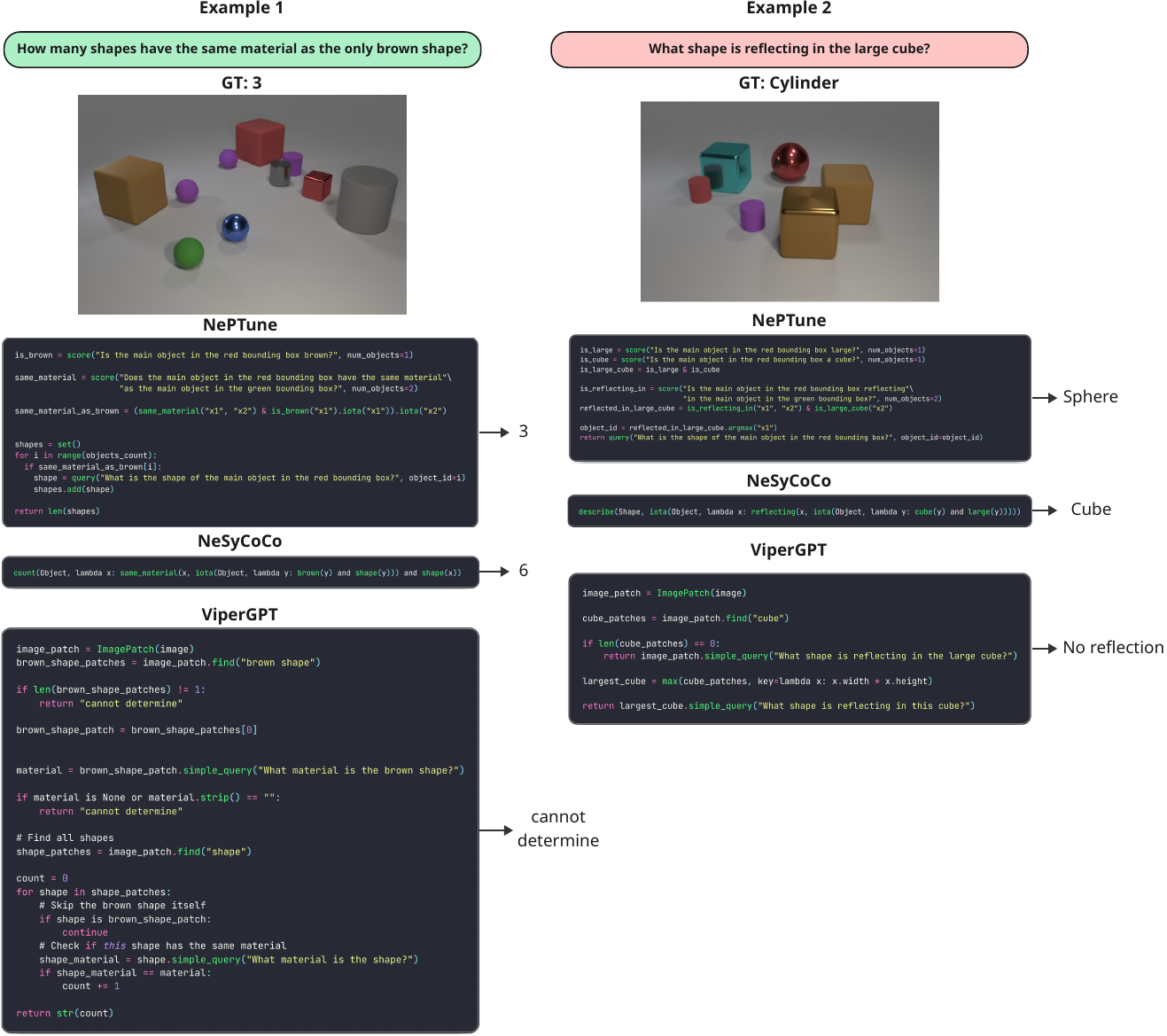}
    \caption{Qualitative examples of NePTune compared to ViperGPT and NeSyCoCo on CLEVR-Humans.}
    \label{fig:qualitative-clevr}
\end{figure}

Examples of NePTune on CLEVR-Humans are shown in Figure~\ref{fig:qualitative-clevr}. 
\textbf{Example 1} illustrates a case where reasoning over meta-concepts such as distinct shapes is required. 
NeSyCoCo is unable to resolve the query due to this limitation to declarative, while ViperGPT struggles to identify the correct anchor object because it does not perform global reasoning across the object set. 
NePTune, by contrast, is able to provide the correct interpretation.   \textbf{Example 2} presents a more complex scenario where reflection reasoning and calibration are critical. 
Both the sphere and the cylinder are reflected in the cyan cube, with the cylinder’s reflection being more prominent and thus the more likely answer. 
NePTune does not fully capture this subtle distinction and produces an incorrect response. 
The competing baselines, however, fail for different and more fundamental reasons: VLM score calibration issues lead to incorrect selection, and ViperGPT relies on bounding box size comparisons that ignore perspective, which causes it to treat equally sized cubes as different. 
In this case, the code selects the shiny brown cube in the foreground and concludes that there is no reflection. 
Although NePTune also errs, this example highlights the inherent difficulty of fine-grained reflection reasoning and the challenges posed by subtle visual cues. 

\subsection{Failure Example.} 
\begin{figure}[h!]
    \centering
    
    \includegraphics[width=1\linewidth]{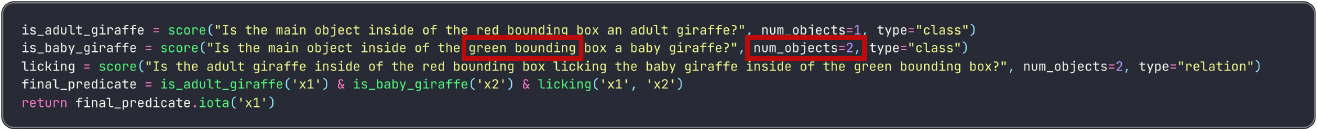}
    \caption{Qualitative example of wrong NePTune program generation. Mistakes are highlighted with red boxes.}
    \label{fig:qualitative-failure}
\end{figure}

Example shown in Figure~\ref{fig:qualitative-failure} illustrates an error caused by an incorrect LLM-generated program in NePTune. Two issues occur simultaneously:  
\begin{enumerate}
    \item \textbf{Incorrect predicate arity:} The program invokes the function \texttt{is\_baby\_giraffe} with \texttt{num\_object=2} instead of \texttt{1}. This leads to a dimensional mismatch when the resulting score tensor is composed with variable \texttt{x2}, causing execution failure.  
    \item \textbf{Incorrect bounding box reference:} The program refers to the bounding box with the color \emph{green} instead of \emph{red}, which is the intended single-object reference.  
\end{enumerate}

While errors such as incorrect bounding box colors can be alleviated to a certain extent with regex-based matching, the dimensional inconsistency in predicate scoring is unrecoverable without regeneration.

\section{VLM Concept Analysis}
\label{app:atomic}

\subsection{Concept-Level Performance}
\begin{table}[h]
\centering
\caption{F1 Scores for atomic concept grounding on CLEVR using visual prompting.}
\setlength{\tabcolsep}{1mm}
\begin{tabular}{lccc}
\toprule
\textbf{Concept}             & \textbf{Ovis1.6} & \textbf{Qwen2VL} & \textbf{InternVL2.5} \\
\toprule
blue                & \textbf{0.972}         & 0.806         & 0.986         \\ 
brown               & \textbf{0.989}         & 0.921         & \textbf{0.989}         \\ 
cube                & 0.915         & 0.952         & \textbf{0.991}         \\ 
cyan                & 0.961         & 0.805         & \textbf{1.000}         \\ 
cylinder            & 0.964         & 0.960         & \textbf{1.000}         \\ 
gray                & 0.958         & 0.911         & \textbf{0.980}         \\ 
green               & \textbf{0.989}         & 0.838         & 0.988         \\ 
large               & 0.956         & 0.975         & \textbf{1.000}         \\ 
metal               & 0.961         & 0.965         & \textbf{0.997}         \\ 
purple              & 0.988         & 0.927         & \textbf{1.000}         \\ 
red                 & 0.973         & 0.829         & \textbf{1.000}         \\ 
rubber              & 0.980         & 0.924         & \textbf{1.000}         \\ 
small               & 0.976         & 0.868         & \textbf{1.000}         \\ 
sphere              & 0.980         & 0.980         & \textbf{0.996}         \\  
yellow              & 0.974         & 0.871         & \textbf{0.982}         \\
\midrule
behind              & 0.836         & 0.877         & \textbf{0.868}         \\ 
front               & \textbf{0.890}         & 0.850         & 0.850         \\ 
left                & \textbf{0.919}         & 0.885         & 0.943         \\ 
right               & \textbf{0.934}         & 0.863         & 0.930         \\ 
same\_color         & \textbf{0.982}         & 0.957         & 0.840         \\ 
same\_shape         & 0.862         & 0.855         & \textbf{0.936}         \\ 
same\_size          & 0.794         & 0.604         & \textbf{0.951}         \\ 
\midrule
\midrule
\textbf{Macro F1}   & 0.943         & 0.874         & \textbf{0.964}         \\ 
\textbf{Micro F1}   & 0.903         & 0.866         & \textbf{0.945}         \\ 
\bottomrule
\end{tabular}

\label{tab:atomic_merged_detailed}

\end{table}

An analysis of the atomic concept grounding performance on the CLEVR dataset shown in Table~\ref{tab:atomic_merged_detailed} reveals clear patterns of weakness, primarily centered around analogical and complex relational concepts.

\subsubsection{Analogical Relations} The most significant challenge for the VLMs concept is abstract, analogical comparisons. The concept \emph{same size} is a clear example, yielding the lowest F1 score in the entire table for Qwen2VL-7B at 0.604. Ovis1.6-9B also struggles significantly with it, scoring just {0.794}. While InternVL2.5-8B performs better, \emph{same color} is a notable weak point for it at 0.840. This indicates a systemic difficulty in performing comparative judgments with visual prompting compared to simple identification. 

\subsubsection{3D Spatial Relations are a Close Second} The next most difficult category is 3D spatial relationships. Concepts like \emph{behind} and \emph{front} consistently score lower than basic attributes across all models. For instance, the F1 scores for \emph{behind} are 0.836 (Ovis1.6), 0.877 (Qwen2VL), and 0.868 (InternVL2.5). This demonstrates a weaker understanding of object relations in three-dimensional space compared to intrinsic properties like shape or material.

\subsubsection{Color Confusion} Certain colors also pose challenges, particularly for the Qwen2VL model. The color \emph{cyan} shows a notable performance dip for Qwen2VL (0.805). Further analysis suggests this is due to confusion with visually similar colors like \emph{blue} (0.806) and \emph{green} (0.838). This pattern highlights that while models can identify common colors well, they are less robust with more specific shades and can struggle to differentiate them.

\subsubsection{Real-World Image Challenges in Visual Genome}

Transitioning from the synthetic CLEVR environment to the complex, real-world images of the Visual Genome~\citep{krishna2016visualgenomeconnectinglanguage} dataset reveals a notable shift in performance dynamics. We evaluate our models on atomic queries generated from the cleaned VG scene graphs in GQA~\citep{gqa} using these templates:
\begin{itemize}
    \item \textbf{Relation Prompt:} \\
    Is the \{class$_1$/object$_1$\} in the red bounding box \{class$_2$/object$_2$\} the \{relation\} in the green bounding box?

    \item \textbf{Class Prompt:} \\
    Is the object \{inside of/in\} the red bounding box \{article\} \{class\}?

    \item \textbf{Attribute Prompt:} \\
    Is the \{class\} in the red bounding box \{attribute\}?
\end{itemize}

As shown in Table~\ref{tab:vg_f1_comparison}, a key challenge emerges in fundamental object identification. While the models remain highly proficient at identifying spatial relations and attributes, their performance in the "Class/Object" category is consistently lower than CLEVR. Ovis1.6-9B drops to an F1-score of 0.804, with InternVL2.5-8B and Qwen2VL-7B at 0.818 and 0.833, respectively. This suggests that the visual complexity, clutter, and vast number of fine-grained categories in real-world images make the foundational task of object recognition a more significant hurdle than in the controlled CLEVR environment.

\begin{table}[h]
\centering
\caption{Top 20 common concepts F1-Score comparison on real-world scene graphs.}

\setlength{\tabcolsep}{1mm}
\small
\begin{tabular}{l c c c}
\toprule
\textbf{Concept} & \textbf{Ovis1.6-9B} & \textbf{Qwen2VL-7B} & \textbf{InternVL2.5} \\
\toprule
right of & 0.940 & \textbf{0.971} & 0.962 \\
left of  & 0.951 & 0.970 & \textbf{0.980} \\
man             & 0.906 & \textbf{0.914} & 0.892 \\
shirt           & 0.840 & 0.893 & \textbf{0.963} \\
person          & 0.821 & 0.792 & \textbf{0.824} \\
window          & 0.857 & 0.894 & \textbf{0.957} \\
pole            & 0.842 & 0.900 & \textbf{0.927} \\
building        & 0.621 & 0.839 & \textbf{0.889} \\
tree            & 0.750 & 0.833 & \textbf{0.839} \\
wall            & 0.815 & \textbf{0.857} & 0.800 \\
on              & 0.636 & \textbf{0.727} & 0.696 \\
sign            & 0.800 & 0.815 & \textbf{0.833} \\
car             & 0.870 & 0.960 & \textbf{1.000} \\
hand            & \textbf{0.929} & 0.929 & 0.923 \\
in              & 0.444 & 0.421 & \textbf{0.500} \\
white           & 0.750 & \textbf{0.889} & 0.846 \\
near            & \textbf{0.923} & 0.880 & \textbf{0.923} \\
sky             & 0.500 & 0.667 & \textbf{0.800} \\
ear             & 0.818 & 0.800 & \textbf{0.909} \\
woman           & \textbf{0.952} & \textbf{0.952} & \textbf{0.952} \\

\bottomrule
\end{tabular}
\label{tab:vg_f1_comparison}
\end{table}

\subsection{End-to-End Performance}

These findings align with the paper's broader conclusion that errors in the final reasoning pipeline often originate from the VLM's poor performance on these specific types of relational and analogical concepts. Based on the results of this analysis, in addition to the \textit{score} perceptual interface, we generate a set of the most common spatial predicates, such as \textit{left}, \textit{right}, \textit{behind}, \textit{front}, etc., using the position and depth estimation of bounding boxes for the sake of efficiency and accuracy.

\begin{table}[h]
\centering
\caption{Comparison of model accuracies across various categories. Maximum values for each NePTune and end-to-end are bolded. Changes compared to normal models are marked with colors and arrows.}
\resizebox{\linewidth}{!}{%
\begin{tabular}{lccc|ccc}
\toprule
\textbf{Metric} & \textbf{Qwen2VL} & \textbf{Ovis1.6} & \textbf{InternVL2.5} & \textbf{Qwen-NeSy} & \textbf{Ovis-Nesy} & \textbf{Intern-Nesy} \\
\toprule
Final Accuracy (\%) & \textbf{93.55} & 85.00 & 90.25 & 82.55 (\textcolor{red}{$\downarrow$ 11.00}) & 88.80 (\textcolor{green}{$\uparrow$ 3.80}) & \textbf{92.65} (\textcolor{green}{$\uparrow$ 2.40}) \\
\midrule
Exist (\%) & \textbf{98.21} & 86.74 & 87.10 & 84.59 (\textcolor{red}{$\downarrow$ 13.62}) & 88.53 (\textcolor{green}{$\uparrow$ 1.79}) & \textbf{93.19} (\textcolor{green}{$\uparrow$ 6.09}) \\
Query Attribute (\%) & 93.77 & 90.14 & \textbf{98.26} & 91.30 (\textcolor{red}{$\downarrow$ 2.47}) & 96.23 (\textcolor{green}{$\uparrow$ 6.09}) & \textbf{96.81} (\textcolor{red}{$\downarrow$ 1.45}) \\
Compare Attribute (\%) & \textbf{99.44} & 90.28 & 98.61 & 78.33 (\textcolor{red}{$\downarrow$ 21.11}) & 88.33 (\textcolor{red}{$\downarrow$ 1.95}) & \textbf{91.94} (\textcolor{red}{$\downarrow$ 6.67}) \\
Count (\%) & \textbf{87.50} & 75.40 & 74.60 & 70.16 (\textcolor{red}{$\downarrow$ 17.34}) & 78.83 (\textcolor{green}{$\uparrow$ 3.43}) & \textbf{87.10} (\textcolor{green}{$\uparrow$ 12.50}) \\

Compare Number (\%) & 90.29 & 78.29 & \textbf{90.86} & 88.57 (\textcolor{red}{$\downarrow$ 1.72}) & 89.14 (\textcolor{green}{$\uparrow$ 10.85}) & \textbf{92.57} (\textcolor{green}{$\uparrow$ 1.71}) \\
\bottomrule
\end{tabular}
}

\label{tab:clevr-multi}
\end{table}

The results in Table \ref{tab:clevr-multi} clearly show that NePTune enhances the compositional reasoning abilities of strong base VLMs such as  Ovis1.6 and InternVL2.5. The most significant gains are on quantitative compositional tasks, with accuracy on counting questions improving by over 12\% for InternVL2.5 and counting comparison questions by nearly 11\% for Ovis1.6. This highlights the value of our structured, programmatic approach for tasks that end-to-end models find challenging. Conversely, our framework degrades the performance of the already accurate model Qwen2VL. A deeper look reveals that the primary source of this degradation is the analogical questions, which are the most apparent in the "Compare Attribute" task, where performance drops by 21\%. This aligns perfectly with our findings in Experiment 1, which showed that this specific VLM struggles with analogical reasoning (e.g., \textit{same color}, \textit{same shape}). This demonstrates both the power of our framework when paired with a reliable grounding VLM and its sensitivity to the underlying VLM's weaknesses, as errors in perception are propagated by the logical executor.

\section{Experimental Setup}
All experiments were conducted on a server running Ubuntu OS, equipped with an AMD EPYC 7413 24-core CPU, 700GB of system RAM, and an NVIDIA A6000 GPU (48GB). Unless specified otherwise, our core models included {DeepSeekV3}~\citep{deepseekai2025deepseekv3technicalreport} as the backbone LLM, {GroundingDINO-B} for object detection, and {DepthAnythingV2}~\citep{depth_anything_v2} for depth estimation. We utilized the Hugging Face Transformers library~\citep{transformers} for VLM interaction and the {InternVL-2.5} codebase for finetuning\footnote{https://internvl.readthedocs.io/en/latest/internvl2.5/finetune.html}. To supplement the LLM's output, we also used {spaCy}~\citep{spacy} Named Entity Recognition to extract keywords from the query.

\section{Datasets}
\label{app:datasets}

Here, we provide a detailed breakdown of the datasets used to evaluate the NePTune framework across a range of visual reasoning challenges.

\subsection{CLEVR}
\label{subsec:clevr}
We evaluate core compositional reasoning using the \textbf{CLEVR (Compositional Language and Elementary Visual Reasoning)} dataset \citep{johnson2017clevr}. It contains 3D-rendered synthetic images of simple objects and is used for the task of \textbf{Visual Question Answering (VQA)}, where the model must answer complex, programmatically generated questions about object attributes, counts, and relations. We measure performance using \textbf{Accuracy (\%)} and, for answers requiring semantic equivalence matching (e.g., matching ``2'' with ``two''). For CLEVR dataset benchmarks where a deterministic evaluation did not capture the correctness of the answer, such as matching "2" with "two", we use GPT-4o as an AI judge for evaluation. The prompt used for this evaluation is shown in Figure~\ref{fig:vqa-eval-prompt}. Since the CLVER evaluation set was large (700K), we used the first 2000 samples for evaluating our method.

\subsubsection{CLEVR Query Categories}
\label{subsec:clevr_categories}
Our fine-grained analysis on CLEVR, presented in Table~\ref{tab:clevr}, breaks down performance across the following five distinct types of reasoning skills using the programs in the dataset:

\begin{description}
    \item[Exist] These questions test for the presence of an object with specific properties, requiring a ``Yes/No'' answer. For example, \textit{``Is there a large green cube behind the small red sphere?''}

    \item[Query Attribute] These questions ask for a specific property (e.g., color, material) of a uniquely identified object, such as, \textit{``What color is the small shiny cylinder?''}

    \item[Compare Attribute] This category involves ``Yes/No'' questions that compare a single property between two objects. For example, \textit{``Does the large sphere have the same material as the small cube?''}

    \item[Count] These questions require the model to return the total number of objects matching a description, such as, \textit{``How many red metallic objects are there?''}

    \item[Compare Number] These questions involve comparing the quantities of two different sets of objects, also resulting in a ``Yes/No'' answer. For example, \textit{``Are there more spheres than cubes?''}
\end{description}
\subsection{CLEVR-Humans}
\label{subsec:clevr_humans}
The \textbf{CLEVR-Humans} dataset \citep{johnson2017inferringexecutingprogramsvisual} allows us to evaluate performance on more natural language. It uses the same synthetic images as CLEVR but features more complex and linguistically diverse questions written by humans. Similar to CLVER, performance is measured by Accuracy (\%). 

\subsection{CLEVR Extensions}
\label{subsec:clevr_extensions}
To probe a wider range of reasoning skills, we use a collection of compositional challenges from Hsu et al., 2024, built on the CLEVR environment \citep{hsu2024s}. Each task tests a unique aspect of complex reasoning:

\begin{description}
    \item[CLEVR-Ref] 
    This task tests referring expression grounding, where the model must identify a target object based on its relationship to other objects. The evaluation is conducted by measuring the IOU value greater than 0.5.
    \begin{quote}
        \textit{Example: ``There is a sphere that is front the gray cylinder, find the small cylinder that is left of it.''}
    \end{quote}

    \item[CLEVR-Puzzles] 
    This benchmark evaluates the model's ability to solve visual constraint satisfaction problems by finding a set of objects that simultaneously satisfy multiple attribute and relational constraints. 
    \begin{quote}
        \textit{Example: ``Can you find four objects from the image such that: object 1 is a large metal object; object 2 is a metal object; object 3 is a small rubber cylinder; object 4 is a small yellow metal cylinder; object 1 is front object 2; object 1 is behind object 4; object 1 is behind object 3.''}
    \end{quote}

    \item[CLEVR-RPM] 
    This task tests abstract relational reasoning by mimicking Raven's Progressive Matrices. The model is presented with objects in a grid that follow a pattern and must identify a candidate object from the scene that correctly completes that pattern.
    \begin{quote}
        \textit{Example: ``There are 9 objects, ordered in a 3x3 grid: row 1 col 1 is a small rubber object; row 1 col 2 is a small rubber object; row 1 col 3 is a large rubber object; row 2 col 1 is a small metal object; row 2 col 2 is a small metal object; row 2 col 3 is a large metal object; row 3 col 1 is a small metal object; row 3 col 2 is a small metal object; I am missing one object at row 2 col 2. Can you find an object in the scene that can fit there?''}
    \end{quote}
\end{description}

\subsection{RefCOCO-Adversarial (Ref-Adv)}
\label{subsec:refcoco_adv}
To test performance in real-world scenarios, we use the \textbf{RefCOCO-Adversarial} benchmark \citep{refadv}. This dataset consists of real-world images with visually similar distractor objects, making the task of Referring Expression Grounding particularly challenging. REG requires the model to locate the specific object corresponding to a text description, and we evaluate the model using {Grounding Accuracy (\%)}, where a prediction is considered correct if the Intersection over Union (IoU) with the ground-truth bounding box is 0.5 or greater.

\subsection{Ref-GTA}
\label{subsec:ref_gta}
To measure generalization under a significant domain shift, we use the \textbf{Ref-GTA} benchmark \citep{refgta}. This is a REG benchmark where images are sourced from a photo-realistic game simulation, creating an out-of-distribution challenge for models pre-trained on natural images. Performance is measured by Grounding Accuracy (\%), defined by an Intersection over Union (IoU) threshold of 0.5 between the predicted and ground-truth bounding boxes.

\section{Verification}
\label{app:verification}
\begin{figure}[h!]
    \centering
\begin{promptbox}
You're an image analyst designed to check if the highlighted objects in the image meets the query description, and which one is more likely to meet the query description.

The query is: "\{query\}"

Please check the highlighted object "0" [in the red bounding box] and "1" [in the green bounding box] in the image and answer the question: Which object is more likely to meet the query description? Your answer should be "0", "1". Answer with one word or phrase.
\end{promptbox}

    \caption{Prompt used for the Pairwise Arbiter verification strategy.}
    \label{fig:verification-prompt}
\end{figure}

To recover from occasional symbolic execution failures, we add a lightweight \emph{verification} stage that operates after reasoning has produced a candidate answer. We explore two simple strategies:
\begin{enumerate}
    \item \textbf{Pairwise Arbiter:} The model is presented with two candidates, the backbone prediction and the symbolic reasoning output, along with the query. It acts as an arbiter, directly judging which option better satisfies the description. Our prompt, similar to the one used by NAVER for this process, is shown in Figure~\ref{fig:verification-prompt}.
    \item \textbf{Confidence Gating:} We compute a softmax distribution over the symbolic executor’s scores with temperature $T$, and use a threshold $\tau$ to decide whether to trust the executor. If $\max(\operatorname{softmax}) < \tau$, we fallback to the backbone prediction; otherwise, we keep the executor’s output. Both $\tau$ and $T$ are tuned on a 500-example held-out set for each backbone.
\end{enumerate}

Table~\ref{tab:verification} summarizes the performance of both light-weight verification strategies across different backbones. Overall, we observe that the \emph{Pairwise Arbiter} tends to outperform the \emph{Confidence Gating} method, highlighting the benefit of leveraging a VLM to arbitrate between backbone and symbolic predictions.
\begin{table}[h!]
\centering
\caption{Post-verification accuracy on Ref-Adv.}
\label{tab:verification}
\small
\begin{tabular}{@{}l l c c c c@{}}
\toprule
\textbf{Backbone VLM} & \textbf{Verification} & \textbf{Accuracy (\%)} & $\boldsymbol{\tau}$ & $\boldsymbol{T}$ & \textbf{Symbolic Share(\%)} \\
\midrule
\multirow{2}{*}{\textbf{Qwen2VL-7B}} 
  & Confidence Gating & 80.91 & 0.70 & 0.40 & 73.59 \\
  & Pairwise Arbiter  & 80.93 & N/A  & N/A  & 80.23 \\
\midrule
\multirow{2}{*}{\textbf{Ovis1.6-9B}} 
  & Confidence Gating & 60.49 & 0.30 & 0.10 & 86.79 \\
  & Pairwise Arbiter  & 63.25 & N/A  & N/A  & 29.73 \\
\midrule
\multirow{2}{*}{\textbf{InternVL2.5-8B}} 
  & Confidence Gating & 76.65 & 0.60 & 0.50 & 71.06 \\
  & Pairwise Arbiter  & 78.08 & N/A  & N/A  & 77.86 \\
\bottomrule
\end{tabular}
\end{table}

\begin{figure}[h!]
    \centering
\begin{promptbox}
You are an automatic answer checker! I will give you a question and answer, and a generated response, and you will tell me if the response is correct given the ground truth answer. If the response is correct, you will say \textless Yes\textgreater, otherwise you will say \textless No\textgreater. Rubber and Matte are the same. Shiny and Metal or Metallic are the same. Square and Cube are the same. Yellow and Golden colors are similar in the images, too. Balls and spheres are similar. Check if the response is correct given the questions and the ground truth answer, deeply thinking about the context. Answer with \textless Yes\textgreater{} or \textless No\textgreater{} and don't mention them other than for the final answer. Important! Ignore the red bounding box reference in the generated response.\\
Question: ``\{Question\}''\\
Ground Truth Answer: ``\{Answer\}''\\
Generated Response: ``\{Prediction\}''
\end{promptbox}

    \caption{Visual Question Answering judge LLM prompt.}
    \label{fig:vqa-eval-prompt}
\end{figure}

\section{Hyperparameters}
\label{app:hyperparams}
The fine-tuning hyperparameters for our experiments are detailed in Table~\ref{tab:consolidated_hyperparams}. These settings are based on the configurations defined in our training script.

During our experiments, we found that applying LoRA fine-tuning to the vision backbone, in addition to the language model, was highly effective for domain adaptation. To illustrate, without any neuro-symbolic fine-tuning, the base NePTune framework achieved \textbf{56.49\%} accuracy on Ref-GTA. In contrast, a standard fine-tuning approach on the base VLM only reached \textbf{4.19\%} accuracy. This performance gap underscores the effectiveness of our neuro-symbolic method and justifies fine-tuning the vision components to achieve optimal results in novel environments.

\begin{table}[h]
\caption{Hyperparameters for Fine-Tuning Experiments}
\centering
\setlength{\tabcolsep}{1mm}
\small
\begin{tabular}{l c c c}
\toprule
\multirow{2}{*}{\textbf{Hyperparameter}} & \textbf{Standard} & \multicolumn{2}{c}{\textbf{NePTune Neuro-Symbolic Fine-Tuning}} \\
\cmidrule(lr){2-2} \cmidrule(lr){3-4}
 & \textbf{Fine-Tuning} & \textbf{on RefCOCO-Adv} & \textbf{on Ref-GTA} \\
\midrule
\multicolumn{4}{l}{\textit{\textbf{General Training}}} \\
\midrule
Learning Rate (lr) & $4 \times 10^{-5}$ & $4 \times 10^{-5}$ & $4 \times 10^{-5}$ \\
Batch Size & 1 & 1 & 1 \\
Gradient Accumulation & 4 & 4 & 4 \\
Weight Decay & 0.01 & 0.01 & 0.01 \\
LR Scheduler & Cosine & Cosine & Cosine \\
Warmup Ratio & 0.03 & 0.03 & 0.03 \\
Loss Function & Cross-Entropy & Binary Cross-Entropy & Binary Cross-Entropy \\
\midrule
\multicolumn{4}{l}{\textit{\textbf{LoRA Configuration}}} \\
\midrule
Rank (r) & 16 (LM), 8 (Vision) & 16 (LM) & 16 (LM), 8 (Vision) \\
Alpha ($\alpha$) & 32 (LM), 16 (Vision) & 32 (LM) & 32 (LM), 16 (Vision) \\
Dropout & 0.05 & 0.05 & 0.05 \\
\bottomrule

\end{tabular}

\label{tab:consolidated_hyperparams}
\end{table}

\end{document}

%% file: math_commands.tex
\usepackage{amsmath,amsfonts,bm}

\def\eqref#1{equation~\ref{#1}}

\def\1{\bm{1}}

\DeclareMathAlphabet{\mathsfit}{\encodingdefault}{\sfdefault}{m}{sl}
\SetMathAlphabet{\mathsfit}{bold}{\encodingdefault}{\sfdefault}{bx}{n}

